\documentclass[USenglish,twocolumn]{article}
\usepackage[utf8]{inputenc}
\usepackage[big,online]{dgruyter}
\usepackage[sort&compress,square,numbers]{natbib}
\usepackage{times}

\begin{document}

  \DOI{10.1515/cdbme-2026-0267}
  \openaccess
  \pagenumbering{gobble}

\title{Activity Recognition from Smart Insole Sensor Data Using a Circular Dilated CNN}
\runningtitle{Activity Recognition from Smart Insole Sensor Data Using a Circular Dilated CNN}

\author*[1]{Yanhua Zhao}

\runningauthor{Y.Zhao et al.}

\affil[1]{\protect\raggedright 

KIS*MED (AI Systems in Medicine), Technische Universität Darmstadt, 64283 Darmstadt, Germany,
e-mail: yanhua@yanhua-zhao.de}

\abstract{Smart insoles equipped with pressure sensors, accelerometers, and gyroscopes offer a non-intrusive way of monitoring human gait and posture. We present an activity classification system based on a circular dilated convolutional neural network (CDCNN) that processes multi-modal time-series data from such insoles. The model operates on 160-frame windows with 24 channels (18 pressure, 3 accelerometer, 3 gyroscope axes), achieving 86.42\% test accuracy in a subject-independent evaluation on a four-class task (Standing, Walking, Sitting, Tandem). Permutation feature importance reveals the distribution of contributions made by various sensors to the classification process. The proposed approach has potential applications in areas such as smart insole design, health monitoring, and rehabilitation.}


\maketitle

\section{Introduction}
\label{sec:intro}
Wearable sensors embedded in footwear enable continuous monitoring of gait, balance, and posture without restricting natural movement \cite{Lin}. Applications range from fall-risk assessment in older adults \cite{Nyan} to sports performance analysis and rehabilitation. Smart insoles typically integrate pressure mats with inertial measurement units (IMUs), yielding high-dimensional time-series that capture both ground-reaction patterns and foot motion.

Existing work on smart insole systems has explored hand-crafted features and classical machine learning models for gait event detection and activity recognition, for example using time- and frequency-domain pressure statistics combined with support vector machines or random forests \cite{Sazonov,Pan}. More recently, deep learning has become the dominant paradigm for human activity recognition from wearable sensors, with convolutional and recurrent architectures applied to accelerometer and gyroscope streams collected from smartphones and body-worn devices \cite{ref3}. However, many of these approaches focus on inertial data alone, or treat pressure and inertial modalities independently, leaving open the question of how best to fuse all available channels from a smart insole in a unified model.

Beyond wearables, radar \cite{Zhao} and vision-based systems \cite{Zhao2} have also been proposed for activity recognition and gait analysis. Camera setups can capture rich kinematic information but raise privacy concerns, depend on lighting conditions and line-of-sight, and are difficult to deploy outside controlled environments. Radar sensors can operate in low light and offer some through-obstacle capability, yet typically require fixed infrastructure, careful calibration, and sophisticated signal processing, and they still observe subjects from a distance rather than at the point of contact with the ground. In contrast, smart insoles provide subject-specific, viewpoint-invariant, and inherently privacy-preserving measurements that naturally integrate into everyday footwear and can be used continuously in free-living conditions.

In this work we focus on end-to-end learning directly from the raw multi-modal insole signals. We propose a circular dilated 1D convolutional neural network (CDCNN) that treats each time window as a sequence over the feature dimension, applying dilated convolutions with circular padding to exploit temporal dependencies without losing resolution. The model is trained on a dataset of 14,748 labeled windows and evaluated with a held-out test set. We further employ permutation feature importance to quantify the contribution of each sensor channel. It also provides insight into the roles of pressure and inertial measurements.

\begin{figure*}[t]
\centering
\includegraphics[width=0.9\linewidth]{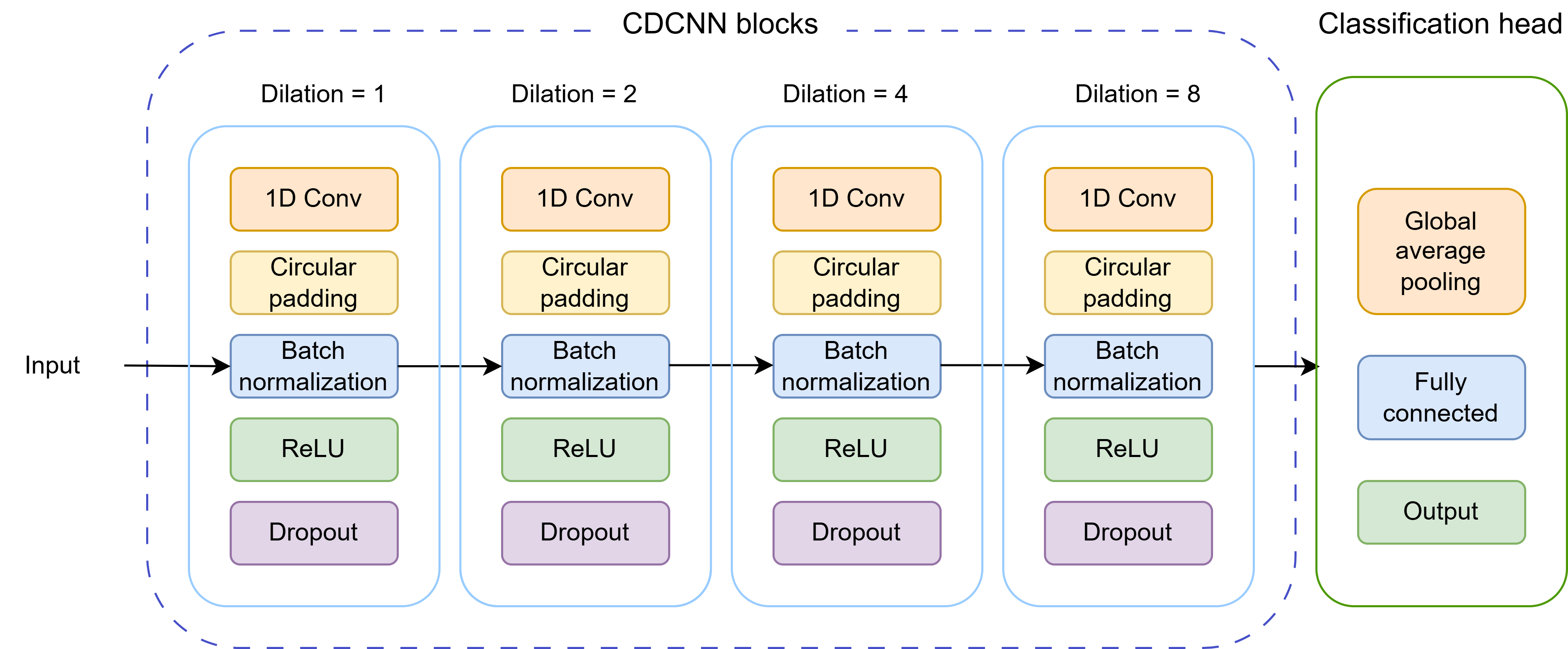}
\caption{CDCNN model structure.}
\label{fig:model}
\end{figure*}

\section{Method}
\label{sec:method}

\subsection{Dataset and Preprocessing}

The dataset is compiled from recordings of multiple subjects performing four activities: Standing, Walking, Sitting, and Tandem \cite{dataset}. Each sample is a fixed-length window of 160 time steps with 24 features per frame:
\begin{itemize}
  \item 18 pressure sensors (pressure\_0 to pressure\_17) sampled per time step
  \item 3-axis accelerometer (accel\_x, accel\_y, accel\_z) in units of g
  \item 3-axis gyroscope (gyro\_x, gyro\_y, gyro\_z)
\end{itemize}

Input matrices are reshaped to $(N, T, F) = (N, 160, 24)$, where $N$ is the batch size, $T$ the number of frames, and $F$ the number of channels. Class distribution is approximately balanced (Standing: 5559; Walking: 5401; Sitting: 5066; Tandem: 5043 samples).

\subsection{CDCNN Architecture}

Compared with standard CNNs or recurrent architectures, the CDCNN offers several advantages for smart insole data. Dilated convolutions expand the effective temporal receptive field exponentially with depth while keeping the number of parameters small \cite{cdcnn}, enabling the model to capture both short-term foot contacts and longer-term gait patterns within a 160-frame window. Circular padding avoids boundary artifacts at the start of each segment, which is important when activities do not align perfectly with window borders. In addition, the purely convolutional design is fully parallelizable across time, leading to lower inference latency than RNN-based models.

Each input sample is a window of $T=160$ time steps with $F=24$ sensor channels (pressure, accelerometer, gyroscope). Features then are sent through a stack of circular dilated convolutional blocks that aggregate information over increasingly longer time scales. A final global average pooling over time produces a compact 64-dimensional representation for each window, which is fed to a linear classifier that outputs logits for the four activity classes. The architecture is shown in Figure \ref{fig:model} and it comprises:

\textbf{Circular dilated convolution blocks.} Each block consists of a 1D convolution with dilation $d = 2^i$ for layer $i$, circular padding to preserve sequence length, batch normalization, and ReLU. Four such blocks use dilations 1, 2, 4, and 8, increasing receptive field without pooling. Hidden channels are set to 64; kernel size is 3. Dropout (0.2) is applied after each block.

\textbf{Classification head.} A global average pooling layer reduces the temporal dimension to 1, followed by a fully connected layer mapping 64 hidden units to the number of classes (4). Cross-entropy loss is minimized with the Adam optimizer (learning rate 0.01). Early stopping with patience 20 on validation accuracy prevents overfitting.

\subsection{Permutation Feature Importance}
To assess which channels contribute most to classification, we apply permutation importance \cite{ref4}. Let $\mathcal{D}_{\text{test}} = \{(x_i, y_i)\}_{i=1}^{N}$ denote the test set and $M$ the trained CDCNN. The baseline accuracy is
\begin{equation}
  A_{\text{base}} = \frac{1}{N} \sum_{i=1}^{N} \mathbf{1}\big[M(x_i) = y_i\big].
\end{equation}
For a given feature (channel) $f$, we construct a permuted dataset $\tilde{\mathcal{D}}^{(f)}$ by randomly shuffling the values of feature $f$ across samples while keeping all other features unchanged. The corresponding accuracy is

\begin{equation}
  A_{f}^{\text{perm}} = \frac{1}{N} \sum_{i=1}^{N} \mathbf{1}\big[M(\tilde{x}_i^{(f)}) = y_i\big],
\end{equation}
where $\tilde{x}_i^{(f)}$ denotes the input with the permuted feature $f$. The permutation importance of feature $f$ is defined as the accuracy drop:
\begin{equation}
  I_f = A_{\text{base}} - A_{f}^{\text{perm}}.
\end{equation}
\begin{figure}[b]
\centering
\includegraphics[width=1\linewidth]{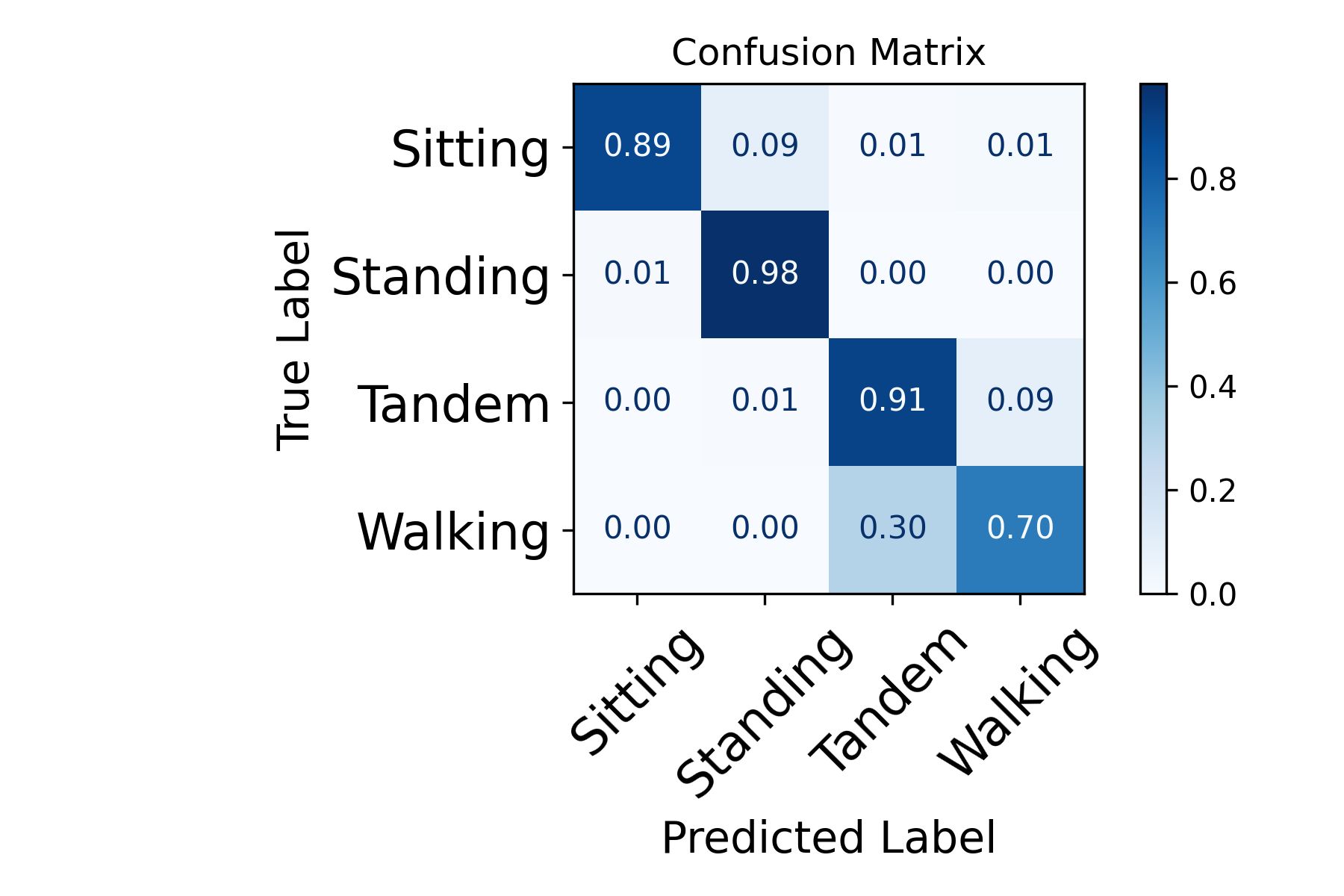}
\caption{Confusion matrix for CDCNN model on the test set.}
\label{fig:cm}
\end{figure}
Larger values of $I_f$ indicate a stronger contribution of feature $f$ to the classifier's performance. This procedure is applied to every channel over the held-out test set.

\begin{table}[h]
\centering
\caption{Number of samples per subject and class.}
\label{tab:subject_class_counts}
\begin{tabular}{lrrrrr}
\toprule
Subject ID & Sitting & Standing & Tandem & Walking & Total \\
\midrule
1  &   0 & 404 &   0 & 426 &  830 \\
10 &   0 & 392 &   0 & 412 &  804 \\
12 &   0 & 430 &   0 & 358 &  788 \\
13 &   0 & 358 &   0 & 360 &  718 \\
14 & 396 & 396 &   0 & 147 &  939 \\
15 & 566 & 398 & 590 & 406 & 1960 \\
16 & 548 & 368 & 215 & 370 & 1501 \\
17 & 504 & 153 & 564 & 418 & 1639 \\
18 & 342 &   0 &   0 &   0 &  342 \\
19 & 592 & 388 & 594 & 240 & 1814 \\
22 & 408 & 390 & 510 & 386 & 1694 \\
23 & 330 & 372 & 546 & 368 & 1616 \\
24 & 230 & 360 & 514 & 366 & 1470 \\
30 & 384 & 382 & 550 & 410 & 1726 \\
31 & 442 & 372 & 420 & 406 & 1640 \\
32 & 324 & 396 & 540 & 328 & 1588 \\
\bottomrule
\end{tabular}
\end{table}

\section{Results}
\label{sec:results}

\subsection{Dataset Overview}

\begin{table}[t]
\centering
\caption{Subject-wise train/validation/test split and sample counts.}
\label{tab:subject_split}
\begin{tabular}{lcr}
\toprule
Split & Subject IDs & Samples \\
\midrule
Train & 1, 10, 12, 14, 15, 18, 19, 23, 30, 31, 32 & 14,047 \\
Val   & 17, 22                                     &  3,333 \\
Test  & 13, 16, 24                                 &  3,689 \\
\bottomrule
\end{tabular}
\end{table}

The dataset contains recordings of multiple subjects performing the four target activities. The dataset is relatively balanced across Standing, Walking, Sitting, and Tandem, which reduces the risk of bias toward any single activity.

Table~\ref{tab:subject_class_counts} details the number of samples per subject and per class, highlighting that some subjects contribute more samples and that the distribution of activities varies slightly across subjects.

To enforce subject independence between training and evaluation, subjects are partitioned into disjoint sets using a fixed random seed. Table~\ref{tab:subject_split} lists the subject IDs assigned to each split and the resulting number of samples.

\subsection{Classification Performance}

Training was run for up to 300 epochs with early stopping on the subject-wise validation set. On the held-out subject-wise test set, the CDCNN achieved a \textbf{test accuracy of 86.42\%}. Confusion matrix on testset is shown in Figure \ref{fig:cm}. It can be observed that the classification performance is generally consistent across all activities, with Walking exhibiting a slightly lower accuracy. In particular, approximately 30 \% of Walking samples in testset are misclassified as Tandem, indicating partial overlap between these two activity patterns.

For comparison, an XGBoost model trained on data of the same subject-wise split reached a higher test accuracy of \textbf{87.83\%}. One likely reason for the performance gap is that XGBoost can exploit rich non-linear interactions between all time-step features in the flattened vector, whereas the CDCNN is constrained by its local convolutional filters and may not be fully tuned to the limited dataset size and cross-subject variability. In addition, the CDCNN hyperparameters (depth, hidden width, regularization) were only coarsely tuned, whereas XGBoost is known to be robust with moderate tuning.

Despite the slightly lower accuracy, the CDCNN offers several potential advantages. First, it operates directly on the sequence structure, enabling extensions to variable-length windows or online streaming without re-flattening. Second, the temporal feature maps of the CDCNN can be inspected or constrained (e.g., via attention or class activation mapping) to provide time-resolved interpretability that is harder to obtain from ensembles of decision trees. The model converges quickly; validation accuracy exceeds 80\% within a few epochs.

\begin{figure}[b]
\centering
\includegraphics[width=1\linewidth]{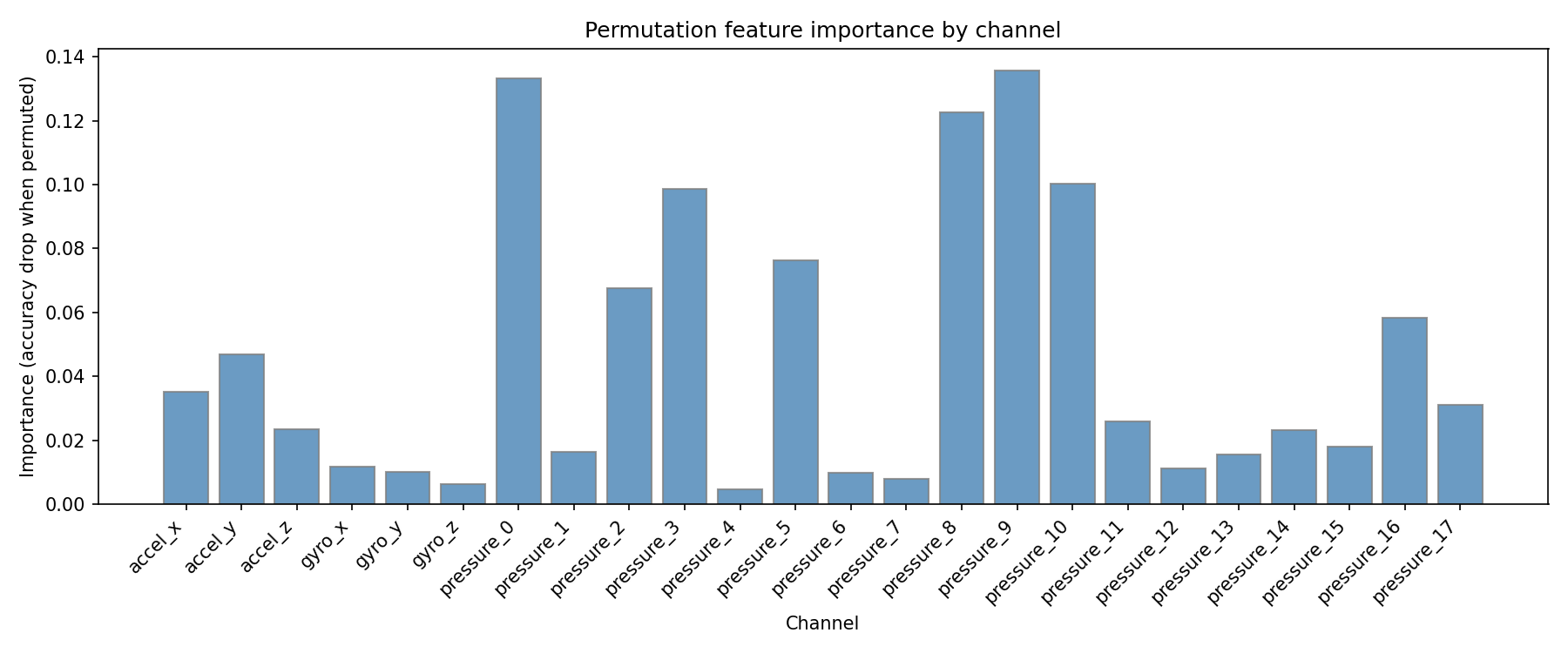}
\caption{\centering Permutation feature importance by channel over the test set.}
\label{fig:feat_importance}
\end{figure}
\begin{figure}[t]
\centering
\includegraphics[width=1\linewidth]{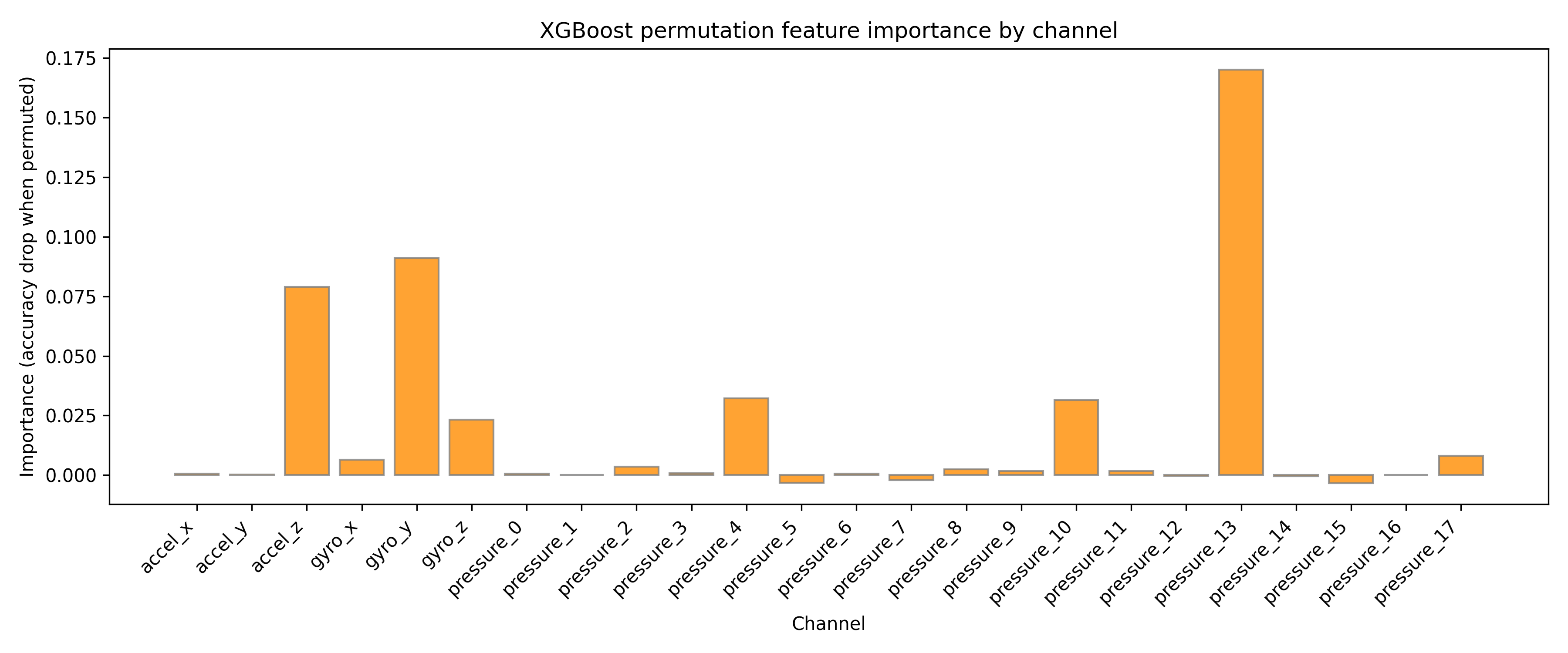}
\caption{\centering Permutation feature importance by channel over the test set for the XGBoost.}
\label{fig:feat_importance_xgb}
\end{figure}

\subsection{Feature Importance}

Permutation importance analysis of the CDCNN shows that pressure channels contribute most strongly to activity classification, with several pressure sensors consistently exhibiting the highest importance scores \ref{fig:feat_importance}. This suggests that pressure information from motion are highly informative for discriminating the four activities. 
For comparison, we also compute permutation importance for the XGBoost on the same subject-wise test set (Fig.~\ref{fig:feat_importance_xgb}), observing different rankings in importance score. This suggests that the two models rely on different subsets of features for classification, with the CDCNN placing greater emphasis on pressure channels, while the XGBoost model assigns relatively higher importance to selected accelerometer and gyroscope channels.

These findings can inform the design of more effective insole-based sensing systems by prioritizing sensor placement and density in regions that capture highly discriminative pressure patterns, while also retaining inertial sensing to complement dynamic motion information.

\section{Conclusion}
\label{sec:conclusion}

We presented a CDCNN-based activity recognition pipeline for smart insole data that fuses pressure and inertial modalities. On a challenging subject-independent split (no subject overlap between train, validation, and test sets), the model achieves 86.42\% test accuracy with a simple, interpretable architecture and performs competitively with an XGBoost model (87.83\%). Permutation feature importance indicates that which sensor channel contribute more to model decision making. Based on this, the distribution of sensors on the insole can be optimised effectively.

\section{Outlook}
\label{sec:outlook}

Future work will address several directions. First, large subject-independent evaluation (leave-one-subject-out) can be conducted to assess generalization of model to new users. Second, we would like to extend the model to recognize finer activities (e.g., stair ascent/descent, turning) and continuous gait events (stance, swing phases), to increase clinical utility. Finally, lightweight model variants and pruning could be considered to enable on-device inference on resource-constrained insoles.


\end{document}